\documentclass[letterpaper]{article} % DO NOT CHANGE THIS
\usepackage{aaai23}  % DO NOT CHANGE THIS
\usepackage{times}  % DO NOT CHANGE THIS
\usepackage{helvet}  % DO NOT CHANGE THIS
\usepackage{courier}  % DO NOT CHANGE THIS
\usepackage[hyphens]{url}  % DO NOT CHANGE THIS
\usepackage{graphicx} % DO NOT CHANGE THIS
\usepackage{natbib}  % DO NOT CHANGE THIS AND DO NOT ADD ANY OPTIONS TO IT
\usepackage{caption} % DO NOT CHANGE THIS AND DO NOT ADD ANY OPTIONS TO IT
\usepackage{algorithm}
\usepackage{algorithmic}

\usepackage{newfloat}
\usepackage{listings}
\DeclareCaptionStyle{ruled}{labelfont=normalfont,labelsep=colon,strut=off} % DO NOT CHANGE THIS
\floatstyle{ruled}
\newfloat{listing}{tb}{lst}{}
\floatname{listing}{Listing}
\title{Sentiment Analysis Across Multiple African Languages: A Current Benchmark}
\author{Saurav K. Aryal$^\dag$\equalcontrib, Howard Prioleau\equalcontrib, and Surakshya Aryal}

\affiliations{
    EECS, Howard University,\\
    Washington, DC 20059, USA\\
    $^\dag$E-mail: saurav.aryal@howard.edu\\
    https://howard.edu/
}

\usepackage{bibentry}
\begin{document}

\maketitle

\begin{abstract}
Sentiment analysis is a fundamental and valuable task in NLP. However, due to limitations in data and technological availability, research into sentiment analysis of African languages has been fragmented and lacking. With the recent release of the AfriSenti-SemEval Shared Task 12, hosted as a part of The 17th International Workshop on Semantic Evaluation, an annotated sentiment analysis of 14 African languages was made available. We benchmarked and compared current state-of-art transformer models across 12 languages and compared the performance of training one-model-per-language versus single-model-all-languages. We also evaluated the performance of standard multilingual models and their ability to learn and transfer cross-lingual representation from non-African to African languages. Our results show that despite work in low resource modeling, more data still produces better models on a per-language basis. Models explicitly developed for African languages outperform other models on all tasks. Additionally, no one-model-fits-all solution exists for a per-language evaluation of the models evaluated. Moreover, for some languages with a smaller sample size, a larger multilingual model may perform better than a dedicated per-language model for sentiment classification.
\end{abstract}

\section{Introduction}

Africa is the second-largest and fastest-growing continent with rich natural resources. However, Africa's adverse climate conditions and geopolitics puts it at a disadvantage in development \cite{source1}. After the mass decolonization of Africa, many countries experienced economic crises of varying severity \cite{source2}, which served as a hindrance to development. Most African nations are still faced with challenges of "meeting basic needs such as education, energy, food, potable water supply, efficient healthcare delivery" for a significant portion of their population \cite{source4}. Although most nations "have national policies and strategies that promote science and technology, but their capacity to implement them remains weak" \cite{source6}. Consequently, government and industry are yet to prioritize funding for research and development (R\&D). \cite{source6}.

We can see the effect of less investment on R\&D in terms of research output. During 2000–2004, European Union and the USA produced 38.8\% and 33.6\% of the world publications, respectively. In contrast, Africa produced 1.8\% of the world publications \cite{source5}. During the same period, while the rest of the world produced 817,197 patents, Africa produced 633 patents (less than 0.1\% of the world's inventions) \cite{source5}. There is a significant limitation provided by the infrastructure as well. As of 2009, African countries had an access rate of 5.4\% to the internet despite the global percentage of 23\% \cite{source3}, which is made worse by widespread illiteracy. Even when there is access to the internet, the "bandwidth is often too narrow" \cite{source4}, which limits the population from accessing information and resources available on the internet.
Similarly, there is also a severe and persisting need for more provision for continuing education and training \cite{source4} with the existing education infrastructure. However, in recent days, Africa has been progressing in growing science, technology, and innovation (STI). Africa's share of world publication output more than doubled since 2003 to reach 3\% today \cite{source6}. A vast consumer population has led private sectors to take an interest in the African market \cite{source3}. There has also been increasing foreign funding upon which many medium and small-sized research systems are dependent \cite{source3}. 

Sentiment analysis is a part of Natural Language Processing (NLP) that categorizes emotions behind digitalized texts mainly into positive, negative, and neutral. In a digital world, sentiment analysis plays a significant role in providing social impact. NLP can be used as a political tool to help bridge the "informational gaps between decision-makers and citizens in terms of preferred, and eventually winning, outcome. Oftentimes, citizens express their opinions on social media, and user sentiment analysis on these social media posts can be used effectively by the "governments to grasp collective citizens' preferences towards specific negotiation processes..." \cite{source7}. This approach can help write policies and laws that work to the will and benefit of the citizens. \citeauthor{source7} identify that sentiment analysis can help decision-makers identify different options and help gain additional insights when making decisions and enforcing new laws/policies.
Furthermore, Africa is a lucrative market for mobile eCommerce, given that there is a 90\% mobile penetration rate for a growing population currently at over 1 billion. However, the growth of the domestic technology industry often needs to catch up. In particular, eCommerce companies can use sentiment analysis to investigate customer issues and address them, which can help make the eCommerce market profitable and more significant. 

The following section reviews relevant research on sentiment analysis for African Languages. This paper seeks to help advance and contribute to the NLP and sentiment analysis literature for African languages by evaluating the performance of current state-of-the-art transformer models and methods across 12 African languages. We end with a presentation of our results and a brief discussion of limitations and future work.

\section{Relevant Works}
\label{sec:rel}
There are over 2000 languages spoken across African nations. However, many of these languages are explicitly oral with limited written texts. Moreover, the generational impact of colonialism has devastated African languages' support, preservation, and integration \cite{alexander2009afrikaans}. These factors and more have contributed to a technological space that does not equitably represent African languages and results in limited datasets and corpora available for research using NLP and sentiment analysis for these languages \cite{martinus}. However, with recent advancements in NLP and growing interest in Africa, some current and relevant work, albeit limited, in modeling language representations and sentiment classification will be covered in the following subsections.

\subsection{Multilingual Models}
Some researchers suggest that about 30\% of all current day languages are African-derived Languages \cite{Robinson2003WritingUL}. There have been large multilingual models covering 100 plus languages such as XLM-R\cite{DBLP:journals/corr/abs-1911-02116} less than 5 of the languages which were officially African. This phenomenon is not tied to XLM-R; it applies to almost all multilingual models, except those specifically targeted at African Languages. Thus, this section will focus on the three models that are aiming to change the lack of Large Scale NLP Models for African Languages, which are AfriBERTa\cite{ogueji-etal-2021-small}, AfroXLMR\cite{alabi-etal-2022-adapting}, and AfroLM\cite{dossou2022afrolm}. Each of these models, detailed below, is the top Performing Multilingual model for African Languages based on a literature review and to the best of our knowledge. For future reference, you can find the languages supported by or pre-trained on each of the models in Table \ref{tab:Model_Language_Support}.

\subsubsection{AfriBERTa}
AfriBERTa \cite{ogueji-etal-2021-small} was one of the first of its kind; of Multilingual Models focused on primarily African Languages. It demonstrated that creating high-performing multilingual models trained on only low-resource languages is more than possible. It comprises 11 African languages, and the total training data used amounts to less than 1 gigabyte of text. Compared to mainstream multilingual high resource language models such as XLM-R\cite{DBLP:journals/corr/abs-1911-02116}, which was trained on 2.5 Terabytes of data, AfriBERTa beat XLM-R and mBERT in Named Entity Recognition (NER) and Text Classification Tasks across most African languages.

\subsubsection{AfroXLMR}
AfroXLMR \cite{alabi-etal-2022-adapting} followed AfriBERTa \cite{ogueji-etal-2021-small} in its development but took a different approach than it. Where AfroXLMR followed multilingual adaptive fine-tuning (MAFT) on XLM-R\cite{DBLP:journals/corr/abs-1911-02116} to add support of 17 of the highest resourced African languages and three other languages that are widely spoken on the continent of Africa. To further their modeling, they also removed all vocabulary tokens from the embedding layer that are non-African writing scripts \cite{ogueji-etal-2021-small}. This adaptation allowed them to create a high-performing multilingual African Language model that is 50\% smaller than XLM-R and is competitive when evaluated in NER, news topic classification, and sentiment classification.

\subsubsection{AfroLM}
AfroLM \cite{dossou2022afrolm} is the most recent in the lineage from AfroXLMR \cite{alabi-etal-2022-adapting}, and AfriBERTa \cite{ogueji-etal-2021-small} of top Performing Multilingual model for African Languages. AfroLM provides a unique approach to the problem of Low Resource African Multilingual model problem; they developed and trained their model from scratch utilizing an Active Learning approach to the problem. While active learning is excellent at addressing low-resource problems, it receives minimal attention in NLP since it requires expert annotations and labeling. While BERT performs well, it still leaves much to be desired in low-resource language problems. With AfroLM active learning approach, the authors were able to outperform AfriBERTa \cite{ogueji-etal-2021-small}, AfroXLMR \cite{alabi-etal-2022-adapting}, and XLM-R\cite{DBLP:journals/corr/abs-1911-02116} in downstream tasks such as NER, topic classification, and sentiment classification. They demonstrated that the performance needed for African languages can be found outside BERT-based models and can be discovered in other approaches.

\subsection{AfriSenti-SemEval / NaijaSenti}

%EDIT CITATIONS USING THIS LINK FOR AFRISENTI CITATION GUIDELINES USING THIS \cite{muhammad-EtAl:2022:LREC, yimam-etal-2020-exploring}

Annotated datasets for Sentiment Analysis derived from African Languages are vastly limited. This paucity has vastly impeded the development of this task. While there have been previous initiatives to expand data access and availability, the AfriSenti-SemEval Shared Task 12, hosted as a part of The 17th International Workshop on Semantic Evaluation, is a concentrated effort and shared a collection of Twitter datasets in 14 African languages for sentiment classification \cite{muhammad-EtAl:2022:LREC, yimam-etal-2020-exploring}. At the time of writing, monolingual sentiment annotated datasets of 12 languages are made available. The task is co-created by the creators of NaijaSenti \cite{muhammad2022naijasenti} and expands on NaijaSenti. They provided 13 datasets comprising 12 different languages, each being a dataset and a dataset composed of all the languages. The 12 African Languages covered are Hausa(HA), Yoruba(YO), Igbo(IG), Nigerian Pigdin(PCM), Amharic(AM), Algerian Arabic(DZ), Moroccan Arabic/Darija(MA), Swahili(SW), Kinyarwanda(KR), Twi(TWI), Mozambican Portuguese(PT), and Xitsonga(Mozambique Dialect) (TS). These languages are derived from a diverse range of African Countries Nigeria, Ethiopia, Kenya, Tanzania, Algeria, Rwanda, Ghana, Mozambique, South Africa, and Morocco, all in different regions of Africa. The data was gathered from Twitter composed of 3 sentiment labels positive, negative, and neutral, with some of the tweets being code-mixed. While most of these languages have a limited amount of corpus, to our knowledge, some languages, such as Xitsonga, have labeled sentiment analysis datasets created for the first time.

\section{Methodology}
This section details the Datasets, Pre-Processing, Modeling, and Evaluation for this work.

\subsection{Datasets}
We utilized all thirteen datasets from the AfriSenti-SemEval Task comprising Hausa(HA), Yoruba(YO), Igbo(IG), Nigerian Pidgin (PCM), Amharic(AM), Algerian Arabic(DZ), Moroccan Arabic/Darija(MA), Swahili(SW), Kinyarwanda(KR), Twi(TWI), Mozambican Portuguese(PT), Xitsonga(Mozambique Dialect) (TS), and a combination of all the 12 language datasets for a multilingual task(ALL). With the sourcing of all the datasets coming from Twitter, it allows us to claim that the performance of these models should mirror their use in a real-world setting. The dataset makeup is seen below in Table \ref{tab:dataset_breakdown}.

% \usepackage{graphicx}
\begin{table}[H]
\centering
\resizebox{\columnwidth}{!}{%
\begin{tabular}{|l|l|l|l|l|}
\hline
\textbf{Langs} & \textbf{Neg}   & \textbf{Neu}   & \textbf{Pos}   & \textbf{Total} \\ \hline
HA    & 5467  & 5808  & 5574  & 16849 \\ \hline
YO    & 2315  & 3871  & 4426  & 10612 \\ \hline
IG    & 3070  & 5319  & 3644  & 12033 \\ \hline
PCM   & 4054  & 93    & 2255  & 6402  \\ \hline
AM    & 1936  & 3880  & 1665  & 7481  \\ \hline
DZ    & 1115  & 428   & 522   & 2065  \\ \hline
MA    & 1802  & 2350  & 1925  & 6077  \\ \hline
SW    & 239   & 1340  & 684   & 2263  \\ \hline
KR    & 1433  & 1572  & 1124  & 4129  \\ \hline
TWI   & 1462  & 580   & 1827  & 3869  \\ \hline
PT    & 978   & 2000  & 852   & 3830  \\ \hline
TS    & 356   & 171   & 480   & 1007  \\ \hline
ALL   & 24449 & 27693 & 25196 & 77338 \\ \hline
\end{tabular}%
}
\caption{Sentence Labels of Each Dataset}
\label{tab:dataset_breakdown}
\end{table}

% \usepackage{graphicx}
\begin{table}[H]
\centering
\resizebox{\columnwidth}{!}{%
\begin{tabular}{|l|l|l|l|l|}
\hline
\textbf{Langs} & \textbf{Train} & \textbf{Val}  & \textbf{Test}  & \textbf{Total} \\ \hline
HA   & 12754 & 1418 & 2677  & 16849 \\ \hline
YO   & 7669  & 853  & 2090  & 10612 \\ \hline
IG   & 9172  & 1020 & 1841  & 12033 \\ \hline
PCM  & 4608  & 513  & 1281  & 6402  \\ \hline
AM   & 5385  & 599  & 1497  & 7481  \\ \hline
DZ   & 1485  & 166  & 414   & 2065  \\ \hline
MA   & 5024  & 559  & 494  & 6077  \\ \hline
SW   & 1629  & 181  & 453   & 2263  \\ \hline
KR   & 2971  & 331  & 827   & 4129  \\ \hline
TWI  & 3132  & 349  & 388   & 3869  \\ \hline
PT   & 2756  & 307  & 767   & 3830  \\ \hline
TS   & 723   & 81   & 203   & 1007  \\ \hline
ALL  & 57316 & 6369 & 13653 & 77338 \\ \hline
\end{tabular}%
}
\caption{Sample Sizes by Each Dataset}
\label{tab:sampleSizes}
\end{table}

The dataset is roughly balanced by labels outside of PCM (Nigerian Pigdin), DZ (Algerian Arabic), SW (Swahili), TWI, PT(Mozambican Portuguese), and TS(Xitsonga).

\subsection{Pre-Processing}
At the time of writing and experimentation, the official test set of the datasets had not been made available. We utilized the development set in lieu of the Test Set and performed a 90/10 random stratified split on the official training set for our training and validation sets. The sample size of each split utilized can be seen in Table \ref{tab:sampleSizes}. Then, we cleaned the data by removing the English stop-words, punctuation, and digits from the sentences and denoising the social media text. After cleaning, the cleaned sentences passed through the model-specific tokenizer \cite{wolf2019huggingface}. We set the max sentence token value to 20 using the average number of tokens in a sentence across all languages; the sentences with fewer than 20 tokens were padded with zeros. After tokenization, we finetune and evaluate the chosen models as detailed in the following subsection.

\subsection{Modeling and Evaluation}
This section contains our steps for Per Language and Multilingual Modeling. We selected to train four models including XLM-R \cite{DBLP:journals/corr/abs-1911-02116}, AfriBERTa \cite{ogueji-etal-2021-small}, AfroXLMR \cite{alabi-etal-2022-adapting}, and AfroLM \cite{dossou2022afrolm}.  XLM-R sets a baseline since it is pretrained on the least number of African languages.; it is a large model and provides a baseline for comparison of cross-lingual transfer to African languages. The other models were chosen for their proven track record performance on NLP tasks for disparate African languages.

% \usepackage{graphicx}
\begin{table}[H]
\centering
\resizebox{\columnwidth}{!}{%
\begin{tabular}{|l|l|l|l|l|}
\hline
\textbf{Lang} & \textbf{XLM-R} & \textbf{AfriBERTa} & \textbf{AfroXLMR} & \textbf{AfroLM} \\ \hline
HA   & YES   & YES       & YES      & YES    \\ \hline
YO   & NO    & YES       & YES      & YES    \\ \hline
IG   & NO    & YES       & YES      & YES    \\ \hline
PCM  & NO    & YES       & YES      & YES    \\ \hline
AM   & YES   & YES       & YES      & YES    \\ \hline
DZ   & *YES  & NO        & NO       & NO     \\ \hline
MA   & *YES  & NO        & NO       & NO     \\ \hline
SW   & YES   & YES       & YES      & YES    \\ \hline
KR   & NO    & NO        & YES      & YES    \\ \hline
TWI  & NO    & NO        & YES      & NO     \\ \hline
PT   & *YES  & NO        & NO       & NO     \\ \hline
TS   & NO    & NO        & NO       & NO     \\ \hline
\end{tabular}%
}
\caption{Models Language Support *Means that the model Supports the Language but not the African Variant such as XLM-R supports Portuguese but not explicitly Mozambican Portuguese}
\label{tab:Model_Language_Support}
\end{table}

\subsubsection{Per Language Modeling}
Although some languages share similar origins and roots, modern languages are distinct. To evaluate if a dedicated model better supported the uniqueness of each language, we fine-tuned each of the four models individually on each of the 12 languages. This process resulted in 48 models(4 models x 12 languages); we repeated the process with another 48 on a different seed value to ensure the evaluation was standard and not seed-specific. Each model maintained identical hyperparameters of 5 epochs and a training batch size of 256. The held-out validation set was used while fine-tuning to mitigate over-fitting as much as possible due to the nature of small data sets.

\subsubsection{Multilingual Modeling}
While language-specific models are justifiable, the number of models and sample size required grows linearly with the number of languages. Additionally, multilingual models can capture interdependencies between languages to better represent multiple languages with a single model. Since only 1 dataset is utilized for training each model, we only developed four models and an additional four for evaluation confirmation at a different seed value. Then each model had the identical parameters of 5 epochs and train batch size of 256. Each model maintained identical hyperparameters of 5 epochs and a training batch size of 256. The held-out validation set was used while fine-tuning to mitigate over-fitting.

Model training and inference were performed on a late 2021 Lambda Tensorbook with 16 GB Nvidia GeForce 3080. To enable the reproducibility of our work and help other works in the field, we have made our open-sourced access to our code which involves the entire process from pre-processing to modeling, and evaluation. Readers are encouraged to look at the source code at the URL \textit{http://bit.ly/40yvilf} and reach out to the authors for any further questions they might have regarding the work performed in this paper.

\subsubsection{Evaluation}
Evaluation for Per-Language and Multilingual modeling was done on the test set. Standard, weighted average classification metrics: F1, Precision, Recall, and Accuracy are reported in the following Results section. Since the two different seeds provided nearly identical predictions, we only report the scores from the better of the two models. We further use precision-recall curves to compare the performance of each class for the multilingual model.

\section{Results}

\subsection{Per Language Performance}

% \usepackage{graphicx}
\begin{table}[H]
\centering
\resizebox{\columnwidth}{!}{%
\begin{tabular}{|l|l|l|l|l|l|}
\hline
\textbf{Lang} & \textbf{Model}     & \textbf{F1}  & \textbf{Precision} & \textbf{Recall} & \textbf{Accuracy} \\ \hline
HA   & AfriBERTa & .76 & .76       & .76    & .76      \\ \hline
YO   & AfriBERTa & .71 & .71       & .70    & .70      \\ \hline
IG   & AfriBERTa & .78 & .79       & .78    & .76      \\ \hline
PCM  & AfroXLMR  & .71 & .70       & .72    & .72      \\ \hline
AM   & AfroLM    & .59 & .60       & .61    & .61      \\ \hline
DZ   & AfroXLMR  & .58 & .66       & .63    & .63      \\ \hline
MA   & AfroXLMR  & .77 & .78       & .76    & .76      \\ \hline
SW   & AfriBERTa & .62 & .63       & .64    & .64      \\ \hline
KR   & AfroXLMR  & .57 & .58       & .57    & .57      \\ \hline
TWI  & AfroXLMR  & .58 & .58       & .60    & .60      \\ \hline
PT   & AfroLM    & .53 & .53       & .55    & .55      \\ \hline
TS   & AfriBERTa & .54 & .56       & .58    & .58      \\ \hline
\end{tabular}%
}
\caption{Best Performing Models (Highest Weighted-F1) Per-Language}
\label{tab:perLangPerformance}
\end{table}

Upon comparing Tables \ref{tab:sampleSizes} and \ref{tab:Model_Language_Support} with the average performance of Per-language models' performance in Table \ref{tab:perLangPerformance}, we first observe that the performance is proportional to training size. Furthermore, we see that the models originally pre-trained on more African languages have higher generalized performance on African languages than their counterparts, as seen by the superior performance of AfriBERTa, AfroLM, and AfroXLMR. Results also show that no one-model-fit-all solution works across different languages for per-language modeling. Unsurprisingly, none of the languages performed best using a generic XLM-R model. Out of all the models, XLM-R was most affected by overfitting since the validation error was much lower than the test error. These results further support the necessity of further research into African Languages since the general efforts are insufficient, and specialization may be needed.

\subsection{Multilingual Performance}
% \usepackage{graphicx}
\begin{table}[H]
\centering
\resizebox{\columnwidth}{!}{%
\begin{tabular}{|l|l|l|l|l|}
\hline
                  & \textbf{F1}  & \textbf{Precision} & \textbf{Recall} & \textbf{Accuracy} \\ \hline
XLM-R             & .19          & .13                & .36             & .36               \\ \hline
AfriBERTa         & .66          & .66                & .66             & .66               \\ \hline
\textbf{AfroXLMR} & \textbf{.67} & \textbf{.68}       & \textbf{.67}    & \textbf{.67}      \\ \hline
AfroLM            & .63          & .64                & .63             & .63               \\ \hline
\end{tabular}%
}
\caption{Performance of Models for the Multilingual Dataset}
\label{tab:all_model_performance}
\end{table}

As seen in Table \ref{tab:all_model_performance}, while the three African language-customized models have similar performance capabilities, XLM-R is outperformed by the three African language-customized models in overall performance. This discrepancy further confirms what is seen in Table \ref{tab:perLangPerformance} that a generic XLM-R performs subpar compared to the three African language-customized models at a per-language level. Finally, despite the models' difference in size and approaches, they perform similarly on average, with AfroXLMR slightly outperforming the rest. 

\begin{figure}[H]
\centering
\includegraphics[width=0.9\columnwidth]{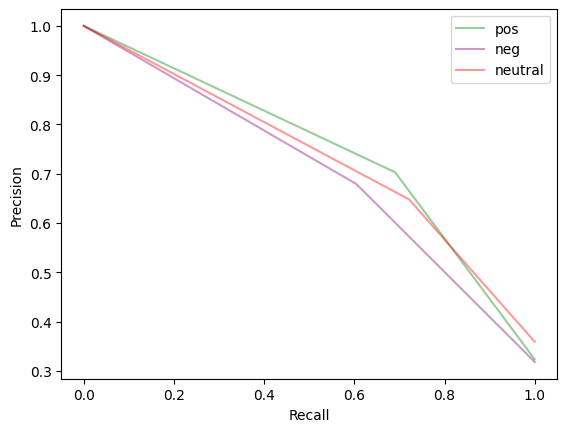} % Reduce the figure size so that it is slightly narrower than the column. Don't use precise values for figure width.This setup will avoid overfull boxes.
\caption{PR curve for best performing model from Table \ref{tab:all_model_performance}}
\label{fig1}
\end{figure}

As seen in Figure \ref{fig1}, we see that for the best performing from Table \ref{tab:all_model_performance} the model performs better in the order: of positive, neutral, and negative. This result is an interesting observation because classification models usually struggle with neutral classes since the difference between positive and negative classes tends to be apparent. Perhaps more research is warranted on the polarity of expression across African languages. 

\section{Conclusion}
With the recent release of the AfriSenti-SemEval Shared Task 12, hosted as a part of The 17th International Workshop on Semantic Evaluation, an annotated sentiment analysis of 14 African languages was made available. We benchmarked and compared current state-of-art transformer models across 12 languages and compared the performance of training one-model-per-language versus single-model-all-languages. We also evaluated the performance of standard multilingual models and their ability to learn and transfer cross-lingual representation from non-African to African languages. Our results show that despite work in low resource modeling, more data still produces better models on a per-language basis. Models explicitly developed for African languages outperform other models on all tasks. Additionally, no one-model-fits-all solution exists for a per-language evaluation of the models evaluated. Moreover, for some languages with a smaller sample size, a larger multilingual model may perform better than a dedicated per-language model for sentiment classification. However, our work is not comprehensive; we implore future researchers and readers to peruse the limitations and future work below.

\section{Limitations \& Future Work} %https://files.slack.com/files-pri/T02FB32SALB-F04KKC948UX/image.png

While sentiment analysis is a fundamental and valuable task in NLP, the potential for abuse of sentiments persists. Mass surveillance, suppression of free speech, and monitoring of dissidents by tyrannical governmental or invasive corporate institutions are potential abuses of these technologies. With the promise of Africa, we can utilize these technologies appropriately and avoid misuse such as the Cambridge Analytica scandal \cite{isaak2018user}.

In terms of limitations of our work, more data would improve performance across the board, so we implore more partnerships with native speakers to expand data access and availability from African academic institutions. In addition, we did not experiment with any hyperparameter optimization which may result in improved performance. Moreover, our approach provides a performance benchmark across multiple languages; the reliance on pre-existing models is limiting and carries pre-existing performance issues and bias. Future work can be to make adaptations suited to African language tokens. We must add that while we had sufficient computing power for this task, this is a privilege not available to all. Additionally, we cleaned the data by removing the English stop-words, punctuation, and digits from the sentences and denoising the social media text. This approach causes a loss of information, such as emoticons or transliterated text which may provide further information for the task at hand. We utilized an average token size for tokenizing each sentence; however, this also limits texts with tokens larger than average. Finally, the authors of this work are not experts or speakers of these languages, so further qualitative analysis of the results is complicated.

\section{Acknowledgments}
The authors would like to acknowledge the support of Dr. Gloria Washington, Director of the Affective Biometrics Lab, and Dr. Legand Burge from the Department of Electrical Engineering and Computer Science at Howard University for providing the computational resources utilized for the project. We are also grateful to Mr. Ishan Baniya for his help in suggesting relevant papers related to sentiment analysis. Finally, we would also like to acknowledge the authors of the AfriSenti-SemEval datasets for sharing and enabling future work in sentiment analysis for African languages. 
\bibliography{citations}

\end{document}